\documentclass[letterpaper, 10 pt, conference]{ieeeconf}  

\IEEEoverridecommandlockouts                              

\usepackage{graphicx}
\usepackage{amsmath}
\usepackage{amssymb}
\usepackage{booktabs}
\usepackage{multirow}
\usepackage{amsfonts}
\usepackage{booktabs}
\usepackage{array}
\usepackage{makecell}
\usepackage{hyperref}

\title{Low-rank Adaptation-based All-Weather Removal \\for Autonomous Navigation}

\author{Sudarshan Rajagopalan and Vishal M. Patel
\thanks{Department of Electrical and Computer Engineering, Johns Hopkins University, USA;
        {\tt\small \{sambasa2, vpatel36\}@jhu.edu}}%
}

\begin{document}

\maketitle
\thispagestyle{empty}
\pagestyle{empty}


\begin{abstract}

All-weather image restoration (AWIR) is crucial for reliable autonomous navigation under adverse weather conditions. AWIR models are trained to address a specific set of weather conditions such as fog, rain, and snow. But this causes them to often struggle with out-of-distribution (OoD) samples or unseen degradations which limits their effectiveness for real-world autonomous navigation. To overcome this issue, existing models must either be retrained or fine-tuned, both of which are inefficient and impractical, with retraining needing access to large datasets, and fine-tuning involving many parameters.  In this paper, we propose using Low-Rank Adaptation (LoRA) to efficiently adapt a pre-trained all-weather model to novel weather restoration tasks. Furthermore, we observe that LoRA lowers the performance of the adapted model on the pre-trained restoration tasks. To address this issue, we introduce a LoRA-based fine-tuning method called LoRA-Align (LoRA-A) which seeks to align the singular vectors of the fine-tuned and pre-trained weight matrices using Singular Value Decomposition (SVD). This alignment helps preserve the model's knowledge of its original tasks while adapting it to unseen tasks. We show that images restored with LoRA and LoRA-A can be effectively used for computer vision tasks in autonomous navigation, such as semantic segmentation and depth estimation. Project page: \href{https://sudraj2002.github.io/loraapage/}{https://sudraj2002.github.io/loraapage/}.

\end{abstract}

\section{Introduction}
\label{sec:intro}
Image restoration under adverse weather conditions is a widely studied problem which is particularly important for autonomous navigation applications. The advent of deep learning has prompted the development of CNN and transformer-based restoration approaches including,~\cite{haze1},~\cite{haze2},~\cite{haze3},~\cite{rain1},~\cite{rain2},~\cite{rain3},~\cite{snow1},~\cite{snow2}, Restormer~\cite{restormer}, MPRNet~\cite{mprnet} and SwinIR~\cite{swinir}. These methods were designed to handle a single degradation at a time, 
making them impractical for real-world scenarios as they require storing multiple sets of weights for different degradations.
\begin{figure}[t]
    \centering
        \includegraphics[height=7cm, width=8cm]{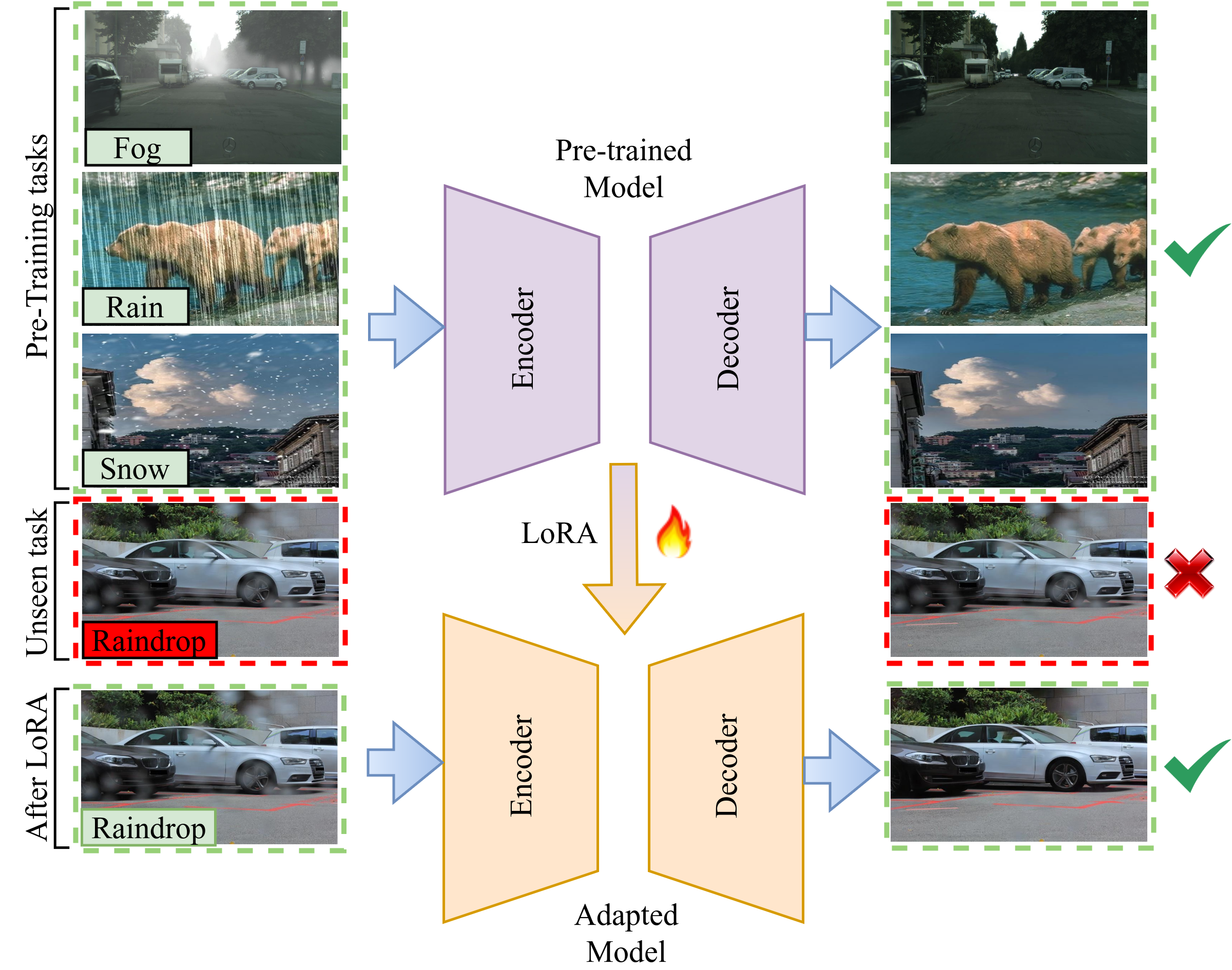}
\vskip-5pt   \caption{The AWIR model shown above is pre-trained for fog, rain and snow, for which it works well. However, it fails for the novel or unseen task of raindrop removal. The performance for raindrop removal improves significantly after parameter efficient adaptation using LoRA.}
    \label{fig:outline}
\end{figure}
To address these challenges, all-weather or all-in-one models such as All-in-one~\cite{nas}, TransWeather~\cite{transw}, Airnet~\cite{airnet}, PromptIR~\cite{promptir} and~\cite{domtrans} have been proposed. These models employ various techniques including multiple encoders, contrastive learning, prompt learning and domain translation, to simultaneously tackle multiple degradations. Despite their proficiency in handling multiple degradations, these networks are fundamentally limited to work for a specific set of training degradations, such as fog, rain and snow. This limitation significantly hinders their deployability for autonomous navigation, as it often necessitates retraining or fine-tuning for any new weather task. As previously mentioned, retraining is not a feasible solution due to its time-consuming nature and the need for access to the original training dataset, which is often large for AWIR tasks. A more practical alternative is to fine-tune the model on a dataset consisting of samples from the new task. However, this is computationally expensive due to the large number of parameters involved.

Recently, Parameter Efficient Fine-Tuning (PEFT) has emerged as an efficient alternative to fine-tuning deep networks. PEFT fine-tunes a pre-trained model for a new task by using only a few learnable parameters.
A pioneering PEFT work, LoRA~\cite{hu2021lora}, showed that the fine-tuning weight updates are often low-rank, enabling a substantial reduction in the number of learnable parameters. They also showed that PEFT methods such as adapters~\cite{adapter1},~\cite{adapter2} and prompt tuning~\cite{promptingnlp},~\cite{chainprompting} under-perform compared to LoRA while having drawbacks such as additional inference latency and difficult optimization. While LoRA has been studied extensively for LLMs~\cite{lora+},~\cite{adalora},~\cite{lorallmsurvey} and large vision models~\cite{cliplora1},~\cite{lvm1},~\cite{explora}, there has been little to almost no focus on its application to the low-level vision task of AWIR. 

In this paper, we propose to employ LoRA to efficiently fine-tune AWIR models for novel restoration tasks. Fig.~\ref{fig:outline} presents an illustration of its working. Images affected by the novel degradation can then be restored by the adapted model and used for autonomous navigation tasks such as depth estimation and semantic segmentation (see Sec.~\ref{subsec:applications}). We also conduct a comprehensive analysis to provide valuable insights on the effective use of LoRA for AWIR tasks. Additionally, we observe that LoRA can lower performance of the adapted model on the original pre-trained tasks which is a disadvantage, especially for autonomous navigation applications. To address this issue, we propose LoRA-Align (LoRA-A) which combines the parameter efficiency of LoRA while preserving the original task performance by seeking alignment of the singular vectors of the pre-trained and fine-tuned weight matrices. This alignment enables the adapted model to be more effective for the original tasks as well, improving its real-world deployability. 

In summary our contributions are as follows:
\begin{enumerate}
    \item We propose LoRA as an efficient fine-tuning technique to adapt pre-trained all-weather restoration models for unseen tasks. We provide valuable insights into the incorporation of LoRA for AWIR models. 
    \item We propose a novel SVD-based alignment method for LoRA called LoRA-Align (LoRA-A) to preserve pre-trained task performance while adapting to novel all-weather restoration tasks.
    \item Additionally, we show that the images restored by LoRA and LoRA-A can be successfully used for downstream tasks such as semantic segmentation and depth estimation, thereby aiding autonomous navigation under adverse weather conditions.
\end{enumerate}

\section{Related Works}
We now discuss relevant research on adverse weather removal, parameter efficient fine-tuning and transfer learning.\\
\noindent {\bf{Adverse weather removal.}}
Early works such as~\cite{early1},~\cite{early2} and~\cite{early3} incorporated degradation-physics to restore images. Subsequently, deep learning based approaches such as SPANet~\cite{rain1},~\cite{rescan},~\cite{rain3},~\cite{icrapaper} and~\cite{rain2} for deraining,~\cite{haze1},~\cite{haze2},~\cite{haze3} and~\cite{haze4} for dehazing, and~\cite{snow100k},~\cite{csd} and~\cite{snow1} for desnowing were proposed. More recent methods such as~\cite{mprnet},~\cite{swinir} and~\cite{restormer} have been proposed to handle multiple degradations.
These methods require storing multiple-sets of weights for each degradation, making them impractical. To address this limitation, AWIR methods have been explored.

\begin{figure*}[t]
    \centering
        \vspace{2pt}\includegraphics[height=5cm, width=17cm]{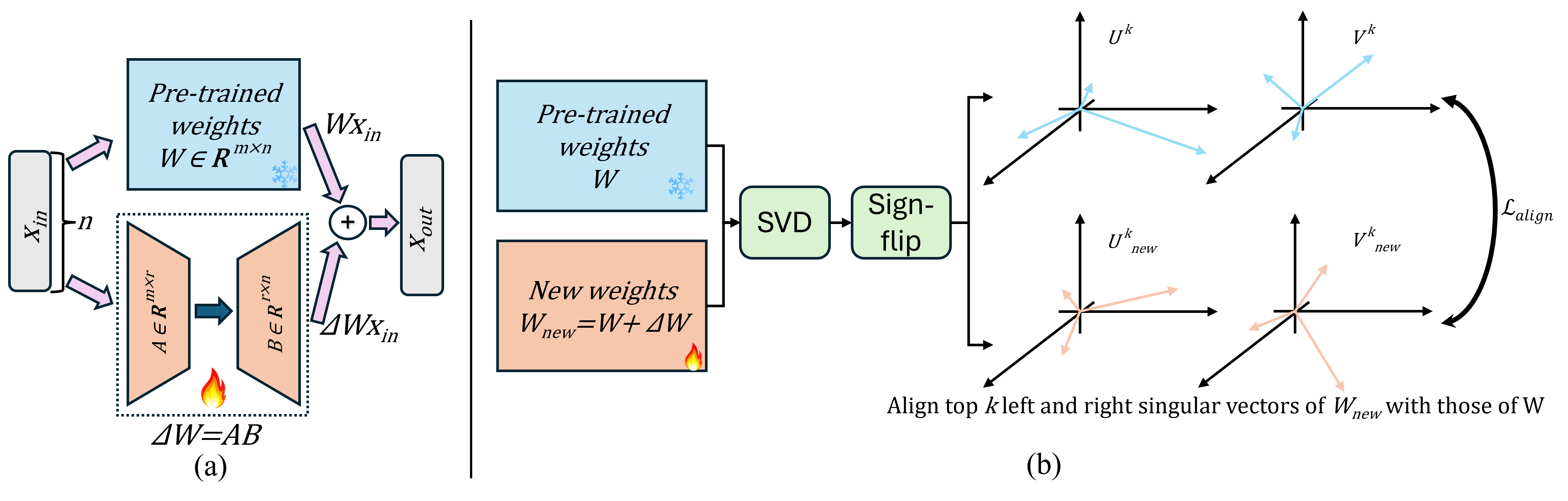}
\vskip-15pt     \caption{Illustration of working of (a) LoRA and (b) LoRA-Align (LoRA-A). LoRA-A uses SVD to obtain the singular vectors of the LoRA-updated weights $W_\text{new}$ and the pre-trained weights $W$. Subsequently, we resolve the sign ambiguity in SVD and obtain the top $k$ sign-corrected left and right singular vectors ($U^k_\text{new}$ and $V^k_\text{new}$) of $W_\text{new}$ and those ($U^k$ and $V^k$) of $W$. The alignment loss $\mathcal{L}_{\text{align}}$ is calculated between $U^k_\text{new}$, $U^k$ and $V^k_\text{new}$, $V^k$ in order to align the singular vectors. This results in LoRA-A retaining performance on the pre-trained tasks. \textit{Graphs are drawn for representation purposes only.}}
    \label{fig:blockdiag}
\end{figure*}

All-in-one~\cite{nas} used neural architecture search with multiple encoders while TransWeather~\cite{transw} employed a unified network with a single encoder for all-weather restoration. Airnet~\cite{airnet} and~\cite{tkmc} used contrastive regularization to learn enhanced degradation representations.~\cite{wgws} fused generic and specific weather features for restoration.~\cite{weatherdiff} utilized a patch-based denoising diffusion model while~\cite{promptir} introduced learnable prompt embeddings for AWIR. Despite these advancements, the deployability of AWIR methods for autonomous navigation is limited by the pre-defined set of training degradations. Adapting to new degradations involves retraining or fine-tuning, both of which are practically inefficient due to the need for learning a large number of parameters.\\
\noindent {\bf{Parameter efficient fine-tuning and transfer learning.}}
The enormous size of LLMs makes full-finetuning for downstream tasks challenging. To address this, parameter efficient fine-tuning (PEFT) and transfer learning methods have been proposed for LLMs. Prompt engineering~\cite{promptingnlp} involves designing task-specific instructions or prompts to adapt a pre-trained model for a new task. Its variants include few-shot prompting~\cite{llama} and chain of thought prompting~\cite{chainprompting}. Visual prompt tuning (VPT)~\cite{vpt} introduced learnable input tokens to achieve a similar functionality for computer vision tasks. SAM~\cite{sam} utilized prompting for various downstream segmentation tasks. Another prominent approach is the use of adapters~\cite{adapter1},~\cite{adapter2}, which are learnable layers added within transformer blocks for novel task adaptation. T2I-adapter~\cite{t2i} leveraged adapters for controllable text to image generation. However, adapters introduce additional latency during inference~\cite{hu2021lora}. A significant advancement in PEFT is LoRA~\cite{hu2021lora}, which proposed a low-rank decomposition of the weight updates of transformer layers. Since its introduction, LoRA has inspired numerous works such as LoRA+~\cite{lora+}, VeRA~\cite{vera}, AdaLoRA~\cite{adalora}, DoRA~\cite{dora}. Despite the growing popularity of LoRA, there has been little focus on its application to the low-level vision task of AWIR. In this work, we aim to demonstrate the potential of LoRA for efficiently adapting pre-trained AWIR models to new tasks, thereby aiding autonomous navigation.

\section{Proposed Methodology}
\label{sec:proposed}
In this section, we present our approach for LoRA-based adaptation of pre-trained AWIR models.
\subsection{Low-rank adaptation}
\label{subsec:lora}
We briefly discuss the working of LoRA~\cite{hu2021lora} for parameter efficient fine-tuning on AWIR models. Fig.~\ref{fig:blockdiag}(a) provides an illustration of how LoRA operates. Consider a dense layer in a pre-trained all-weather model with weights represented by $W \in \mathbb{R}^{m\times n}$. During fine-tuning, the weights are updated to $W + \Delta W$ where $\Delta W\in \mathbb{R}^{m \times n}$ and $W$ is kept frozen. LoRA constraints the learnable parameters in $\Delta W$ to a low-rank decomposition $\Delta W=AB$, where $A\in \mathbb{R}^{m \times r} \text{ and } B\in \mathbb{R}^{r\times n}$. Here, $r$ represents the intrinsic rank of $\Delta W$ and is typically quite small, as demonstrated by~\cite{hu2021lora}. With LoRA, the output $x_\text{out}$ of a dense layer for an input $x_{\text{in}}$ is given by 
\setlength{\belowdisplayskip}{0pt} \setlength{\belowdisplayshortskip}{0pt}
\setlength{\abovedisplayskip}{0pt} \setlength{\abovedisplayshortskip}{0pt}
\begin{equation}
    \label{eq:lora}
     x_\text{out} = W\cdot x_{\text{in}} + \Delta W \cdot x_{\text{in}}.
\end{equation}
Our experiments (see Sec.~\ref{sec:expts}) reveal that low-rank adaptation of pre-trained AWIR models achieves competitive performance on novel restoration tasks, comparable to that of full fine-tuning, while having to learn only a fraction of the parameters. The adapted model can then be used for autonomous navigation tasks, as discussed in Sec.~\ref{subsec:applications}. However, we also observed a fall in performance of the adapted model for the original tasks. To address this issue, we propose an SVD-based alignment method called LoRA-Align (LoRA-A), which can preserve the model's performance on pre-trained tasks while efficiently adapting to new tasks, further enhancing its applicability for autonomous navigation.

\subsection{LoRA-Align}
\label{subsec:LoRA-Align}
We now introduce LoRA-Align (illustrated in Fig.~\ref{fig:blockdiag}(b)), designed to preserve original task performance while adapting a pre-trained AWIR model to novel tasks using LoRA. We accomplish this with the help of SVD. For a dense layer, the SVD of its pre-trained weight matrix $W\in\mathbb{R}^{m \times n}$ is given by $W = U\Sigma V^T$, where $U\in \mathbb{R}^{m \times m}, \Sigma\in \mathbb{R}^{m \times n}, \text{ and } V\in \mathbb{R}^{n \times n}$. The columns of $U$ contain the left-singular vectors of $W$ and the columns of $V$ contain its right-singular vectors. 
$\Sigma$ is a diagonal matrix consisting of the singular values of $W$ arranged in descending order. After one iteration of LoRA, the new weight matrix of the dense layer becomes $W_\text{new}=W + \Delta W$. We decompose the updated weight matrix $W_\text{new}$ using SVD to obtain $W_\text{new} = U_\text{new}\Sigma_\text{new} V_\text{new}^T$ where the dimensions of each of the resulting matrices remain the same as that of $W$. To preserve performance on previous tasks, we align the left and right singular vectors of $W$ with those of $W_\text{new}$, respectively (see Fig.~\ref{fig:blockdiag}), after each iteration of LoRA. This is achieved by selecting the top $k$ singular vectors from $U, U_\text{new}$ and $V, V_\text{new}$, and computing an alignment score between them. However, an important issue arises due to the inherent sign ambiguity in the singular vectors obtained from SVD. This problem is further compounded as we aim to resolve the sign ambiguity across two different weight matrices $W$ and $W_\text{new}$, where  $W_\text{new}$ changes iteratively. To resolve this issue, we propose the following steps:
\begin{enumerate}
    \item Obtain top $k$ singular vectors from $U, V, U_\text{new} \text{ and } V_\text{new}$ as $U^k, V^k, U^k_\text{new} \text{ and } V^k_\text{new}$, respectively.
    \item Compute the ratio $R^i$ = $\frac{||\mathbf{u}^i - \mathbf{u}^i_\text{new}||_2}{||\mathbf{u}^i + \mathbf{u}^i_\text{new}||_2}$ for every left singular vector, $\mathbf{u}^i, i=1, 2,...k$. The rationale behind the definition of $R^i$ is as follows. If the singular vectors $\mathbf{u}^i$ and $\mathbf{u}^i_\text{new}$ are closely related but sign-flipped versions of each other, then $||\mathbf{u}^i + \mathbf{u}^i_\text{new}||_2$ will be very small but $||\mathbf{u}^i - \mathbf{u}^i_\text{new}||_2$ will be a much larger value. Thus, the ratio $R^i$ will become very large in such cases. Conversely, if $\mathbf{u}^i$ and $\mathbf{u}^i_\text{new}$ are closely related but not sign flipped, then the ratio $R^i$ will be quite low. 
    \item If $R^i > T$ (a suitably chosen value), we flip the sign of $\mathbf{u}^i_\text{new}$ and its corresponding right singular vector $\mathbf{v}^i_\text{new}$.
\end{enumerate}
\begin{table*}[t]
    \small
    \setlength{\tabcolsep}{5pt}
    \centering
    \vspace{5pt}
    \caption{Quantitative comparisons of LoRA and LoRA-A with the pre-trained and fine-tuned versions of PromptIR~\cite{promptir} and TransWeather~\cite{transw} for the novel task of raindrop removal. 
    }
    \begin{tabular}{c|c|c|c|c|c|c}
         \multirow{2}{*}{Method}&\multirow{2}{*}{\makecell{Trainable \\parameters (M)}}&\multicolumn{4}{c|}{Pre-trained Tasks}&Novel Task\\
         \cline{3-7}
         &&Fog~\cite{cityfog}&Rain100L~\cite{rain100handl}&Rain100H~\cite{rain100handl}&Snow100k~\cite{snow100k}&Raindrop~\cite{raindrop} \\
         \hline
         PromptIR~\cite{promptir} pre-trained&35.4&28.58/0.972&34.49/0.956&27.25/0.834&33.77/0.936&23.79/0.838 \\
          PromptIR Fine-tuned &35.4&26.66/0.945&25.85/0.822&24.71/0.786&28.52/0.892&29.80/0.904 \\
         PromptIR LoRA &0.55&24.62/0.931&25.51/0.828&24.52/0.781&27.70/0.882&29.63/0.900 \\
         PromptIR LoRA-A &0.55&26.74/0.949&26.88/0.848&24.98/0.790&29.30/0.905&29.35/0.897 \\
         \hline
         TransWeather~\cite{transw} pre-trained&38.31&28.19/0.958&31.58/0.935&26.03/0.805&31.69/0.912&24.01/0.841 \\
         TransWeather Fine-tuned&38.31&18.18/0.864&26.72/0.860&24.10/0.754&26.71/0.855&28.89/0.888\\
        TransWeather LoRA &0.26&20.29/0.890&27.06/0.870&24.04/0.767&27.44/0.867&28.30/0.883\\
         TransWeather LoRA-A&0.26&23.47/0.930&27.70/0.890&25.07/0.793&28.00/0.883&27.84/0.878\\
    \end{tabular}
    \label{tab:raindrop}
\end{table*}

\begin{table}[t]
    \centering
    \small
    \caption{Quantitative comparisons of fine-tuning and LoRA for various pre-training and unseen tasks. The results for the novel task of rain (row 3) are on the Rain100H~\cite{rain100handl} dataset. 
    }
    \begin{tabular}{ >{\centering\arraybackslash}p{1.94cm} | >{\centering\arraybackslash}p{0.8cm} | >
    {\centering\arraybackslash}p{0.8cm} | >{\centering\arraybackslash}p{0.8cm} |>{\centering\arraybackslash}p{0.8cm}}
        Pre-trained tasks&Novel task&Pre-trained&Fine-tuned&LoRA\\
        \hline
        \makecell{Rain,Snow,\\Raindrop}&Fog&\makecell{15.59/\\0.827}&\makecell{23.72/\\0.944}&\makecell{25.41/\\0.947}\\        \makecell{Rain,Raindrop,\\Fog}&Snow&\makecell{23.17/\\0.771}&\makecell{29.85/\\0.892}&\makecell{29.60/\\0.889}\\
        \makecell{Snow,Fog,\\Raindrop}&Rain&\makecell{12.13/\\0.344}&\makecell{25.03/\\0.760}&\makecell{24.78/\\0.750}\\
    \end{tabular}
    \label{tab:cycle}
\end{table}
Since our sign-flipping method would not work correctly if $W_\text{new}$ is significantly different from $W$, we perform the alignment at every LoRA iteration, starting from the first iteration. Subsequently, we calculate the dot product with the sign-corrected vectors and obtain $\mathcal{S}^U_\text{align}$ and $\mathcal{S}^V_\text{align}$ as the alignment scores for left and right singular vectors of $W$ and $W_\text{new}$, respectively.
\begin{subequations}
    \begin{equation}
        \mathcal{S}^U_\text{align} = {U^{k}}^T_\text{new} \cdot U^k, \hspace{2pt}\mathcal{S}^U_\text{align}\in\mathbb{R}^{k\times k}
        \label{eq:scoreu}
    \end{equation}
    \begin{equation}
        \mathcal{S}^V_\text{align} =  {V^{k}}^T_\text{new} \cdot V^k, \hspace{2pt}\mathcal{S}^V_\text{align}\in\mathbb{R}^{k\times k}.
        \label{eq:scorev}
    \end{equation}
\end{subequations}

\noindent These scores are then used to compute an alignment loss, $\mathcal{L}_\text{align}$, for aligning the singular vectors of $W$ and $W_\text{new}$.
\begin{equation}
    \begin{aligned}
        \mathcal{L}_\text{align} = 0.5 \cdot (\text{Mean}(\text{Diag}(I_{k\times k}-\mathcal{S}^U_\text{align})^2) + \\ \text{Mean}(\text{Diag}(I_{k\times k}-\mathcal{S}^V_\text{align})^2)),
        \label{eq:loss}
    \end{aligned}
\end{equation}
\noindent where Mean$(.)$ represents averaging and $\text{Diag}(.)$ selects the diagonal elements from a matrix. $\mathcal{L}_\text{align}$ is computed for each LoRA layer in the model and averaged to obtain the mean alignment loss $\mathcal{L}^{\text{avg}}_\text{align}$. The final training loss $\mathcal{L}_{\text{train}}$ is 
\begin{equation}
    \label{eq: finalloss}
    \mathcal{L}_{\text{train}} = \mathcal{L}_1 (y, \hat{y}) + w_\text{align} \cdot \mathcal{L}^{\text{avg}}_\text{align},
\end{equation}
where $y$ is the ground truth, $\hat{y}$ is the restored image, $\mathcal{L}_1(.)$ is the L1 loss and $w_\text{align}$ is a weighting factor for the alignment loss. Directly computing the alignment loss between $W$ and $\Delta W$ results in unstable gradients due to $\Delta W$ not being full-rank. This is a documented issue with the SVD function in PyTorch~\cite{pytorch}.

\section{Experiments}
\label{sec:expts}
We now demonstrate the effectiveness of LoRA for efficient fine-tuning, and show that LoRA-Align (LoRA-A) helps preserve original task performance during adaptation. We evaluate both methods on two restoration networks: PromptIR~\cite{promptir} and TransWeather~\cite{transw}. Additionally, we conduct comprehensive analysis and ablation studies to provide valuable insights into the application of LoRA and LoRA-A for AWIR tasks. Finally, we present results on autonomous navigation tasks such as semantic segmentation and depth estimation using images restored by the adapted models.
\subsection{Implementation details}
\label{subsec:impl}
For all-weather pre-training, we followed the specifications mentioned in the papers of the restoration networks. For fine-tuning, we used a learning rate of $5\times 10^{-5}$. For LoRA and LoRA-A we used a learning rate of $5\times 10^{-4}$. We employed the AdamW optimizer with $\beta_1=0.9, \beta_2=0.99$, an exponential learning rate scheduler with $\gamma=0.95$ and a batch size of $32$ for all experiments. Our ablations (see Sec.~\ref{subsec:ablt}) reveal that LoRA works well while adapting both Attention and MLP layers of the transformer blocks with a rank of $4$. For LoRA-A, we used $k=16$, $w_\text{align}=100$ and $T=7$ along with the above specifications for LoRA. We used the above parameters for Secs.~\ref{subsec:results} and~\ref{subsec:applications}.

\subsection{Datasets}
\label{subsec:datasets}
We considered four tasks for all our experiments: defogging, deraining, desnowing and raindrop removal. The datasets used were as follows:
\begin{enumerate}
    \item Defogging: City Fog dataset~\cite{cityfog} with $8925$ paired images for training and $4575$ paired images for testing.
    \item Deraining: SRD dataset~\cite{mprnet} with $13,711$ paired images for training. For testing, Rain100H and Rain100L datasets~\cite{rain100handl} with $100$ paired images each.
    \item Desnowing: Snow100k dataset~\cite{snow100k} with $50,000$ paired images for both training and testing.
    \item Raindrop removal: Raindrop~\cite{raindrop} dataset consisting of $861$ images for training and $58$ images for testing.
\end{enumerate}
The specific pre-training and adaptation tasks used in different experiments will be described as needed. For adaptation to a new task, we use $1000$ randomly selected images from its training set (on average) rather than the entire dataset to enable fast adaptation.

\subsection{Results and analysis}
\label{subsec:results}

\begin{figure*}[t]
    \centering
    \setlength{\tabcolsep}{1pt}
    \small
    \vspace{5pt}
    \begin{tabular}{ccccccc}
        &Input&Pre-trained&Fine-tuned& LoRA& LoRA-Align& GT\\
        \rotatebox[origin=c]{90}{City Fog~\cite{cityfog}\hspace{-50pt}}&\includegraphics[height=1.9cm, width=2.5cm]{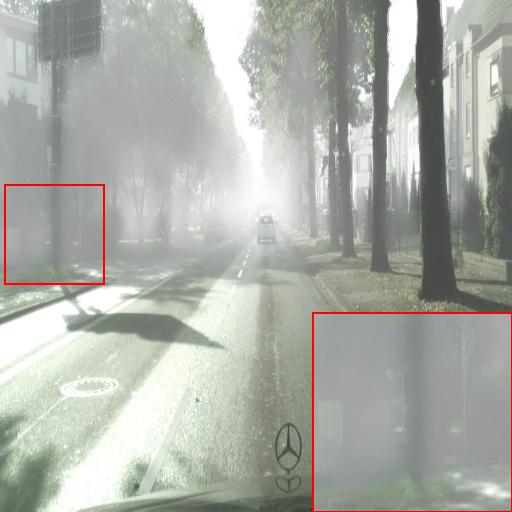}&\includegraphics[height=1.9cm, width=2.5cm]{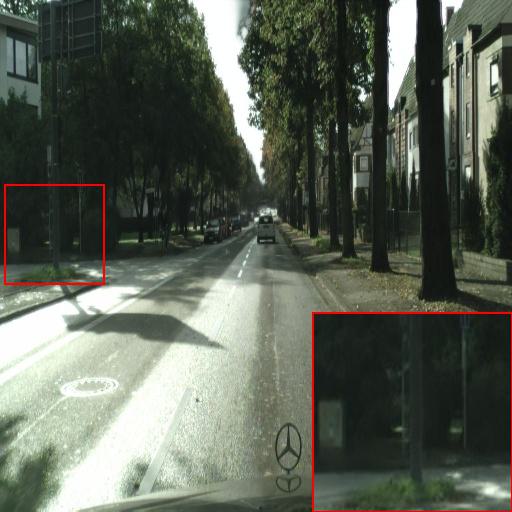}&\includegraphics[height=1.9cm, width=2.5cm]{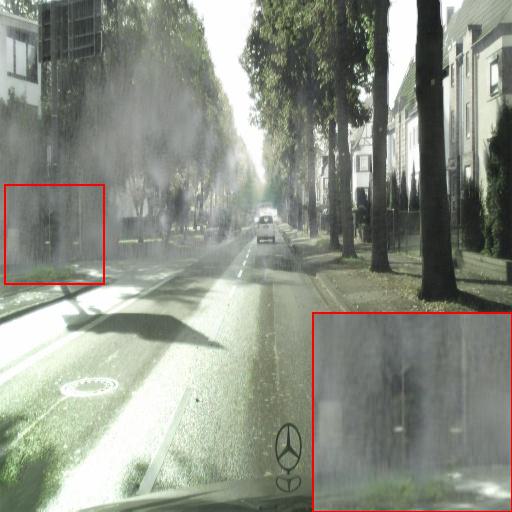}&\includegraphics[height=1.9cm, width=2.5cm]{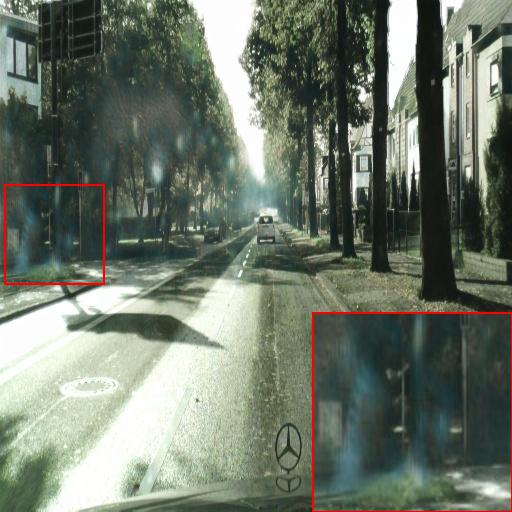}&\includegraphics[height=1.9cm, width=2.5cm]{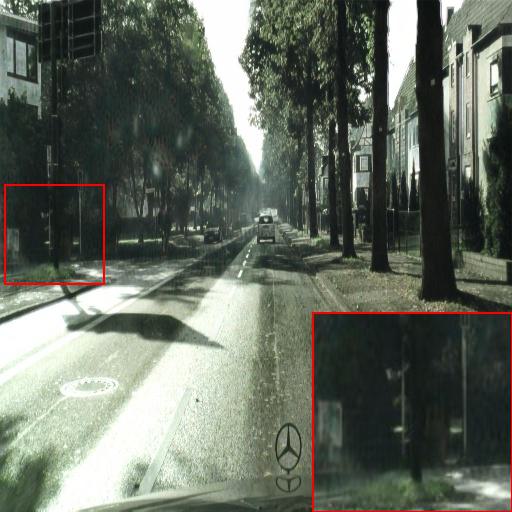}&\includegraphics[height=1.9cm, width=2.5cm]{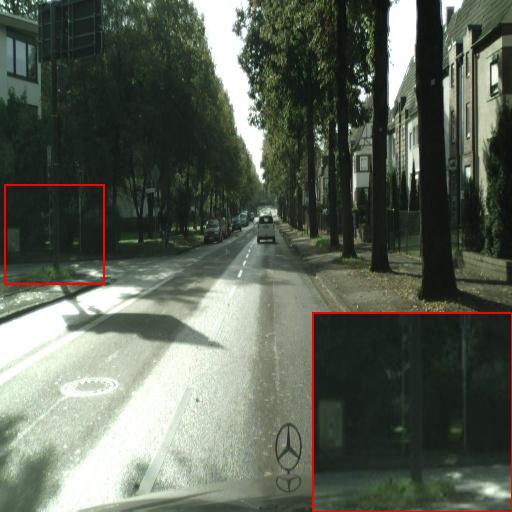}  \\

         \rotatebox{90}{Raindrop~\cite{raindrop}}&\includegraphics[height=1.9cm, width=2.5cm]{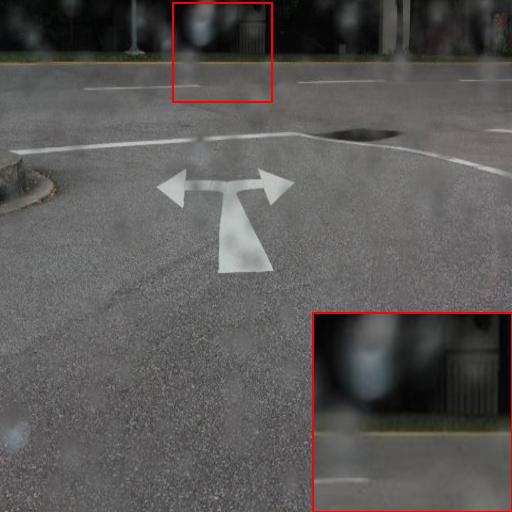}&\includegraphics[height=1.9cm, width=2.5cm]{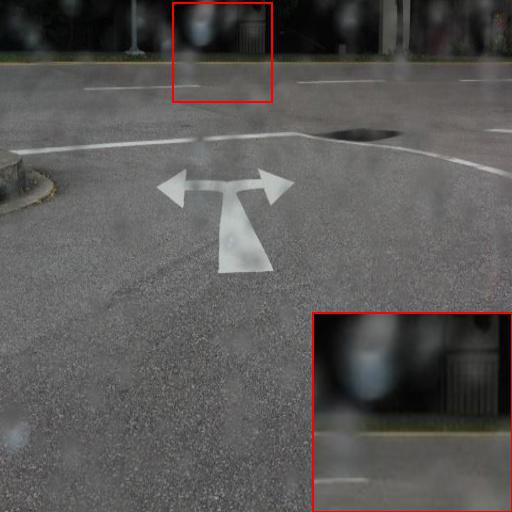}&\includegraphics[height=1.9cm, width=2.5cm]{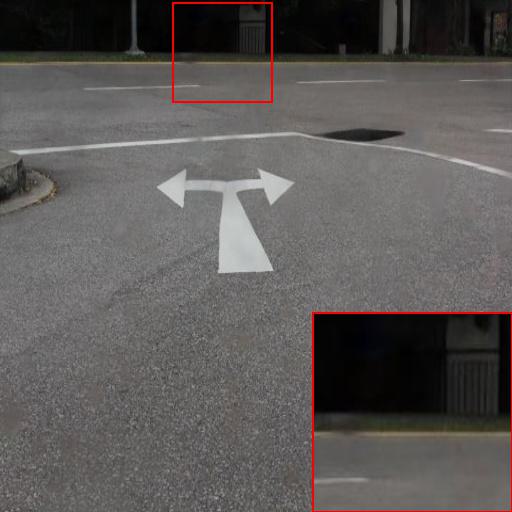}&\includegraphics[height=1.9cm, width=2.5cm]{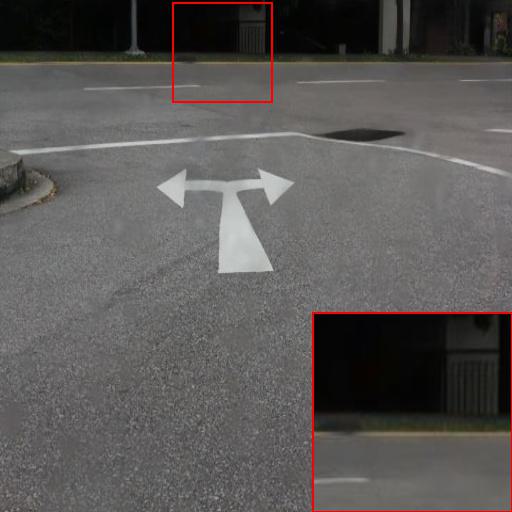}&\includegraphics[height=1.9cm, width=2.5cm]{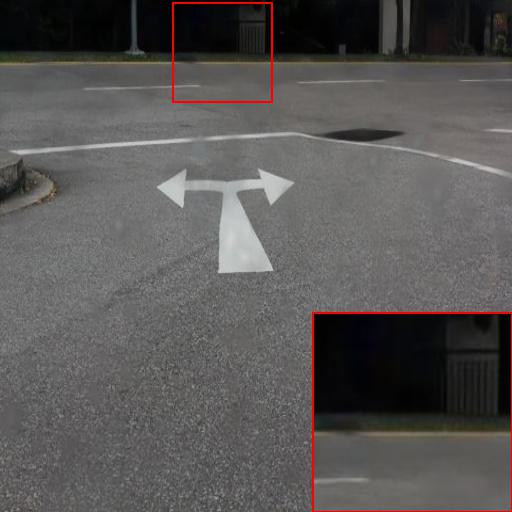}&\includegraphics[height=1.9cm, width=2.5cm]{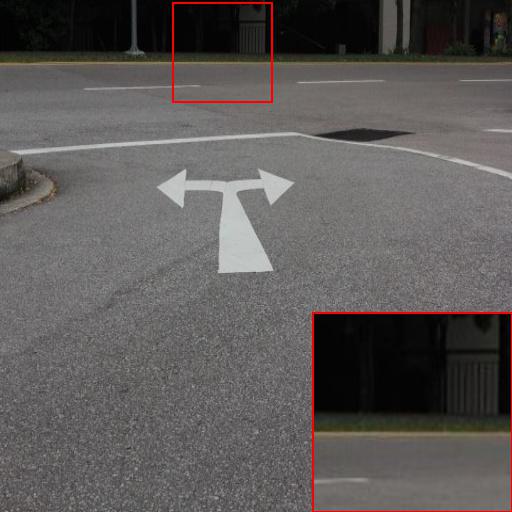}\\
         
    \end{tabular}
\vskip-5pt     \caption{Qualitative comparisons of fine-tuning, LoRA and LoRA-A for the novel task of raindrop removal using PromptIR pre-trained for defogging, deraining and desnowing. Results for one of the pre-training tasks (defogging) is also shown.}
    \label{fig:qual}
\end{figure*}

\noindent \textbf{Performance of LoRA on unseen tasks.} In this experiment, we pre-train PromptIR and TransWeather for defogging, deraining, and desnowing. Then we adapt the models to the novel task of raindrop removal using naive fine-tuning, LoRA and LoRA-A. Table~\ref{tab:raindrop} shows that while the pre-trained models perform well for their original tasks, their performance on the novel task is poor. After fine-tuning and LoRA, the performance of both PromptIR and TransWeather on the novel task improves significantly. Notice that LoRA is on par with the full fine-tuning performance for the novel task of raindrop removal while it needs to learn only a fraction of the parameters ($\sim 1.55\%$ for PromptIR and $\sim 0.68\%$ for TransWeather). The qualitative results for this experiment are shown in Fig.~\ref{fig:qual}. \\ 
\noindent\textbf{Dependence of LoRA on pre-trained tasks.} Our approach of leveraging LoRA for adapting AWIR models to novel restoration tasks does not rely on specific pre-training tasks. To demonstrate this, we pre-train PromptIR on sets of $3$ different tasks chosen from defogging, deraining, desnowing and raindrop removal and use the $4^\text{th}$ unseen task for adaptation. The results of this experiment are presented in Table~\ref{tab:cycle}. Once again, we observe that LoRA achieves competitive performance to full fine-tuning while using only a fraction of the parameters. For the novel task of defogging (see row 1 of Table~\ref{tab:cycle}), LoRA even outperforms fine-tuning, thus, demonstrating its potential for adapting AWIR models.\\
\noindent\textbf{Comparison of LoRA with re-training. }For this experiment, we first pre-train PromptIR~\cite{promptir} for defogging, deraining and desnowing, and adapt it for raindrop removal using LoRA. Next, we re-train a PromptIR model for all four tasks, i.e. defogging, deraining, desnowing and raindrop removal. We then compare the performance of both models for the task of raindrop removal. Table~\ref{tab:retvslora} shows that LoRA with a rank of $4$ achieves nearly the same performance as re-training while using only a fraction of the learnable parameters and trained solely for the novel task of raindrop removal. 
This experiment demonstrates that LoRA-based adaptation can achieve performance close to that of full retraining.\\
\noindent\textbf{Performance of LoRA-Align.} For analysing the performance of LoRA-A, we use the same PromptIR and TransWeather training configurations as for LoRA. LoRA-A has the same parameter efficiency as LoRA while maintaining better performance on the pre-trained tasks (see Table~\ref{tab:raindrop}). Moreover, the table shows that LoRA and LoRA-A achieve near identical performance for the novel task. In the case of PromptIR, for a trade-off of just $0.28$ dB with respect to LoRA on the novel raindrop removal task, LoRA-A remarkably recovers over $1.38$ dB of performance (on the average) on the pre-trained tasks. For TransWeather, LoRA-A preserves $1.35$ dB more performance (on the average) on the pre-trained tasks than LoRA for the raindrop removal task. 
Fig.~\ref{fig:qual} shows qualitative results for one of the pre-training tasks (defogging) and the novel task. Observe that LoRA-A delivers identical performance to LoRA on the novel task while maintaining pre-training task performance. 
Incidentally, when LoRA is trained with rank $2$ so as to reduce its fine-tuning for the novel task (see Table~\ref{tab:ranks}), it yields very similar performance as LoRA-A on the novel task. However, LoRA achieves a PSNR of only $25.70$ dB on the pre-trained task of Rain100L compared to $26.88$ dB of LoRA-A. 

 \begin{table}[t]
    \centering
    \vspace{5pt}
    \caption{Quantitative comparison of LoRA and re-training for the novel task of raindrop removal.}
    \begin{tabular}{c|c|c|c}
         Method&Rank&\#Parameters (M)&PSNR/SSIM  \\
         \hline
         Re-trained&-&35.4M&29.93/0.905\\
         LoRA&4&0.55&29.63/0.900 \\
         LoRA&32&4.4&29.93/0.907  \\
         LoRA&64&8.9&29.91/0.907  \\
    \end{tabular}
    \label{tab:retvslora}
    \end{table}

\subsection{Ablations}
    \label{subsec:ablt}
    \begin{table}[t]
    \centering
    \small
    \caption{Ablations for LoRA on PromptIR to determine the best transformer layers to tune for raindrop removal.}
    \begin{tabular}{c|c|c|c|c}
         Encoder&Decoder&Attention&MLP&PSNR/SSIM  \\
         \hline
         \checkmark&-&\checkmark&\checkmark&28.92/0.885\\
         -&\checkmark&\checkmark&\checkmark&29.28/0.897\\
         \checkmark&\checkmark&\checkmark&-&29.14/0.896\\
         \checkmark&\checkmark&\checkmark&\checkmark&29.63/0.900\\
    \end{tabular}
    \label{tab:placement}
\end{table}

 \begin{table}[t]
    \centering
    \vspace{5pt}
    \caption{Effect of rank of LoRA on the number of learnable parameters (in millions (M)) in PromptIR and its performance on the novel task of raindrop removal.}
    \begin{tabular}{c|c|c|c}
         Method&Rank&\#Parameters (M)&PSNR/SSIM  \\
         \hline
         Fine-tuning&-&35.4&29.80/0.904 \\
         LoRA&2&0.277&29.39/0.899 \\
         LoRA&4&0.554&29.63/0.900 \\
         LoRA&8&1.10&29.67/0.903  \\
         LoRA&16&2.2&29.71/0.905  \\
         LoRA&32&4.4&29.93/0.907  \\
         LoRA&64&8.9&29.91/0.907  \\
    \end{tabular}
    \label{tab:ranks}
\end{table}
    
We conduct ablations on LoRA and LoRA-Align to find the optimal restoration-specific settings for both methods. For all ablations, we use PromptIR pre-trained for defogging, deraining and desnowing, and adapt it to raindrop removal. 

First, we identify the best rank and transformer block layers to apply LoRA. Table~\ref{tab:placement} contains the results obtained by using LoRA on different transformer block layers of the PromptIR model. Specifically, we compare adapting the attention layers, MLP layers, encoder and decoder of PromptIR with a rank of $4$. The results from Table~\ref{tab:placement} indicate that adapting both the attention and MLP layers in the encoder and decoder yields best performance for the novel task of raindrop removal. Table~\ref{tab:ranks} presents the novel task performance for different ranks and the number of learnable parameters involved for each rank. We also provide the performance of full fine-tuning for comparison. It is evident that LoRA achieves competitive performance with full fine-tuning across almost all ranks. In fact, ranks of $32$ and $64$ surpass the performance of full fine-tuning while using considerably fewer parameters. We used a rank of 4 for all our experiments as it provides the best trade-off between performance and parameter efficiency.

Next, we perform ablations on the value of $k$ in LoRA-A where $k$ specifies the number of singular vectors to align for retaining pre-training task performance. We vary $k$ to examine its effect on both the pre-training task and novel task performance. During this experiment, LoRA specifications are used as determined from the previous ablations. Table~\ref{tab:k} shows that increasing the value of $k$ leads to better performance on the pre-training task of deraining on Rain100L. However, the novel task (raindrop removal) performance gradually decreases and the drop becomes quite significant beyond $k=32$. This behaviour is to be expected as we are constraining more singular vectors of the adapted weight matrix ($W_\text{new}$) to align with those of the pre-trained weight matrix $W$. We selected value of $k=16$ as it produces the best trade-off between performance on novel task and pre-trained tasks.

\begin{table}[t]
    \centering
        \caption{Effect of varying $k$ in the LoRA-A framework for the pre-trained task of deraining on Rain100L~\cite{rain100handl} and the novel task of raindrop removal on Raindrop~\cite{raindrop}.}
    \begin{tabular}{c|c|c|c}
    Method&$k$&Rain100L~\cite{rain100handl}&Raindrop~\cite{raindrop}\\
        \hline
        LoRA&-&25.51/0.828&29.63/0.900\\
        LoRA-Align&2&25.96/0.828&29.59/0.901\\
        LoRA-Align&4&26.15/0.832&29.47/0.899\\
        LoRA-Align&8&26.15/0.832&29.52/0.900\\
        LoRA-Align&16&26.88/0.848&29.35/0.897\\
        LoRA-Align&32&27.05/0.854&29.14/0.894\\
        LoRA-Align&64&27.91/0.869&28.14/0.880\\
    \end{tabular}
    \label{tab:k}
    \end{table}


\subsection{Autonomous navigation applications}
\label{subsec:applications}
In this section, we utilize the images restored by PromptIR adapted using LoRA and LoRA-A for two important autonomous navigation tasks: semantic segmentation and depth estimation. For both tasks, we adapt PromptIR pre-trained on defogging, deraining and desnowing for the task of raindrop removal. We use the optimal parameters obtained in Sec.~\ref{subsec:ablt} for LoRA and LoRA-A. Results are presented for the novel task and one of the pre-training tasks (defogging).



\begin{figure}
    \centering
    \setlength{\tabcolsep}{1pt}
    \small
    \begin{tabular}{ccccc}
         &Image&\makecell{Degraded\\ prediction}&LoRA&LoRA-A  \\

         \rotatebox[origin=c]{90}{Novel\hspace{-40pt}}&\includegraphics[height=1.7cm, width=2cm]{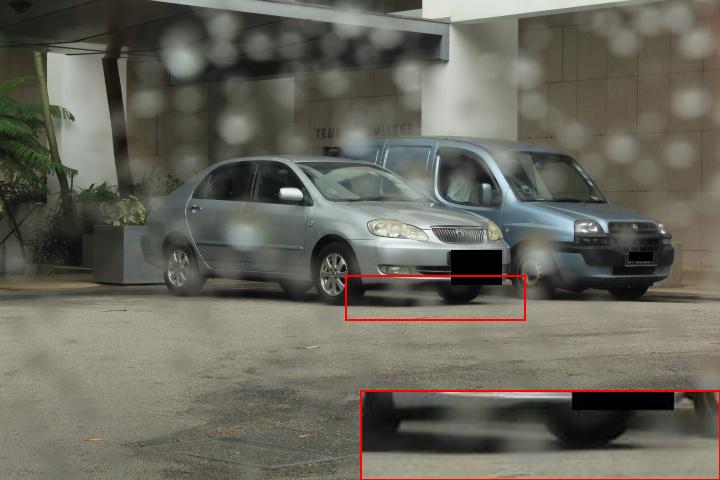}&\includegraphics[height=1.7cm, width=2cm]{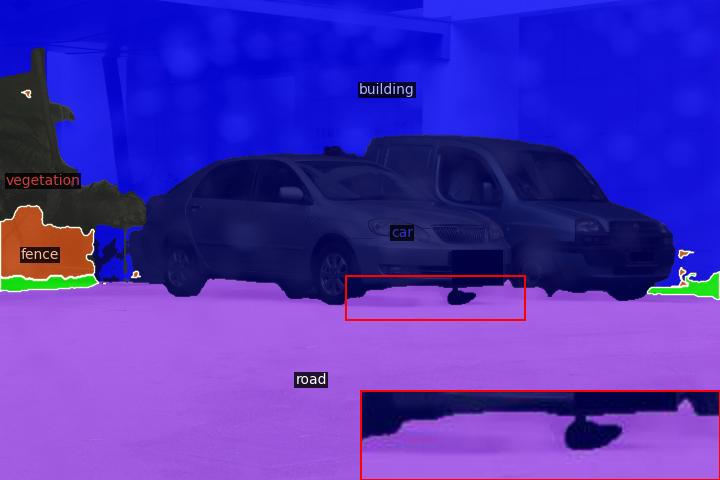}&\includegraphics[height=1.7cm, width=2cm]{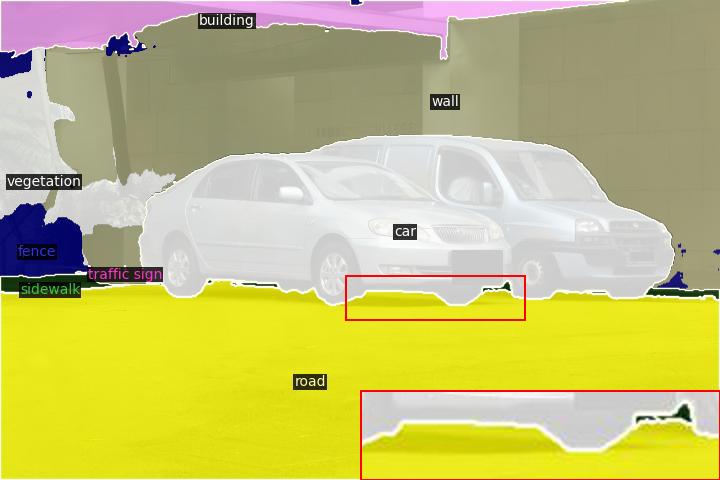}&\includegraphics[height=1.7cm, width=2cm]{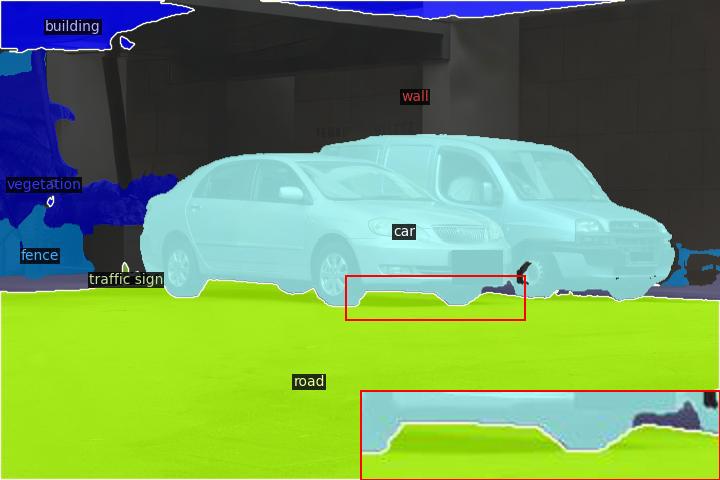}\\

         \rotatebox[origin=c]{90}{Pre-train \hspace{-45pt}}&\includegraphics[height=1.7cm, width=2cm]{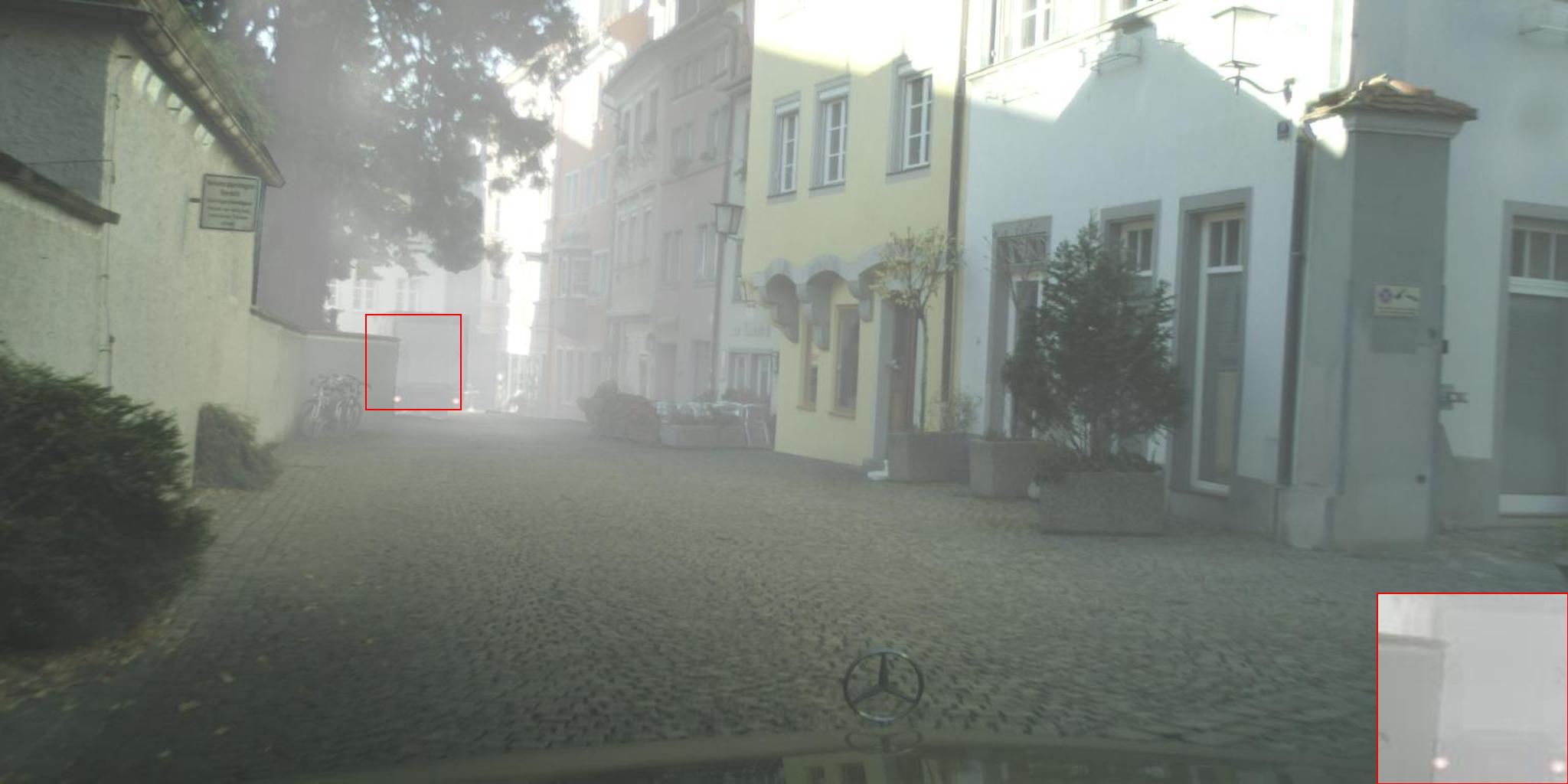}&\includegraphics[height=1.7cm, width=2cm]{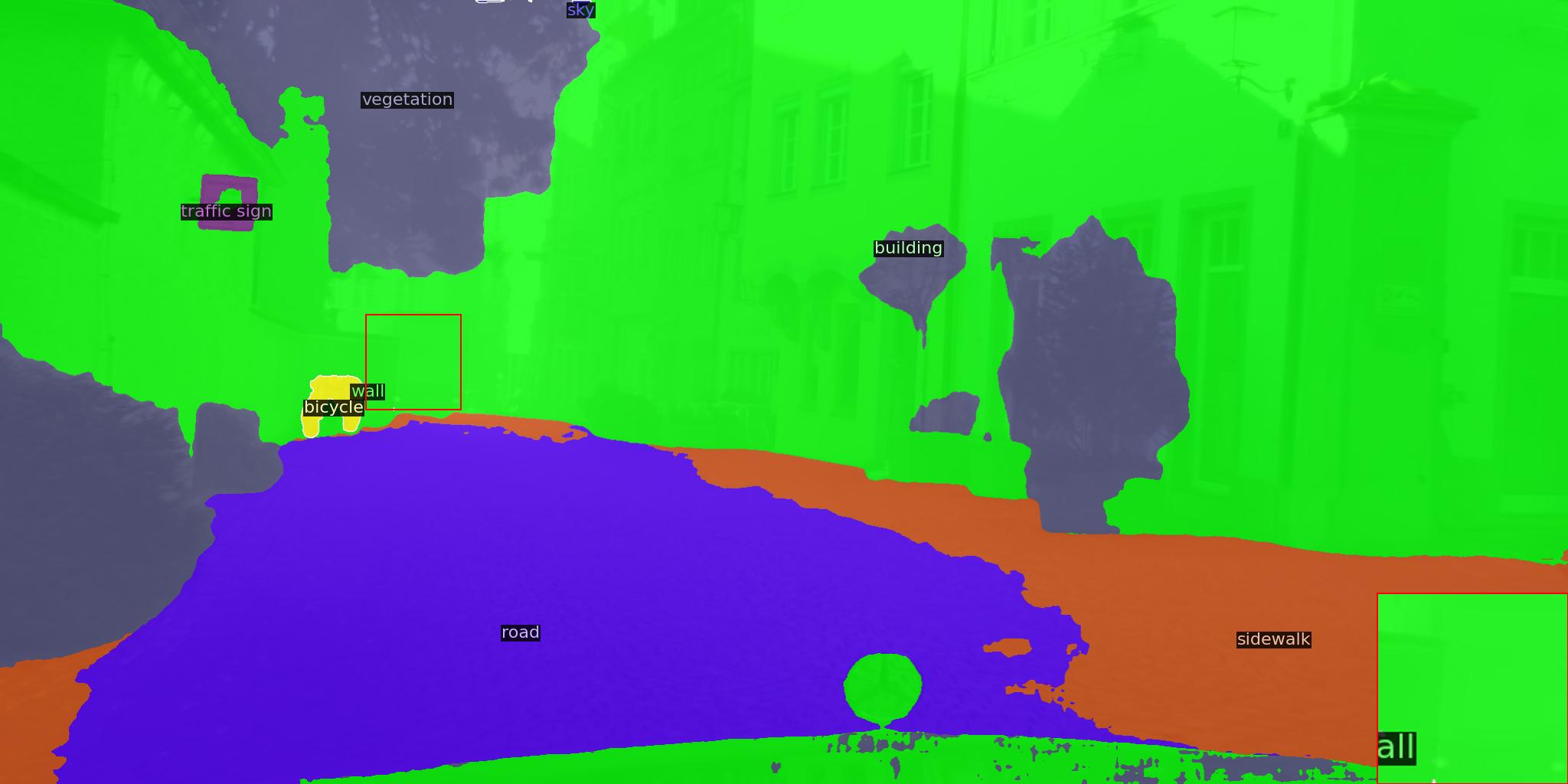}&\includegraphics[height=1.7cm, width=2cm]{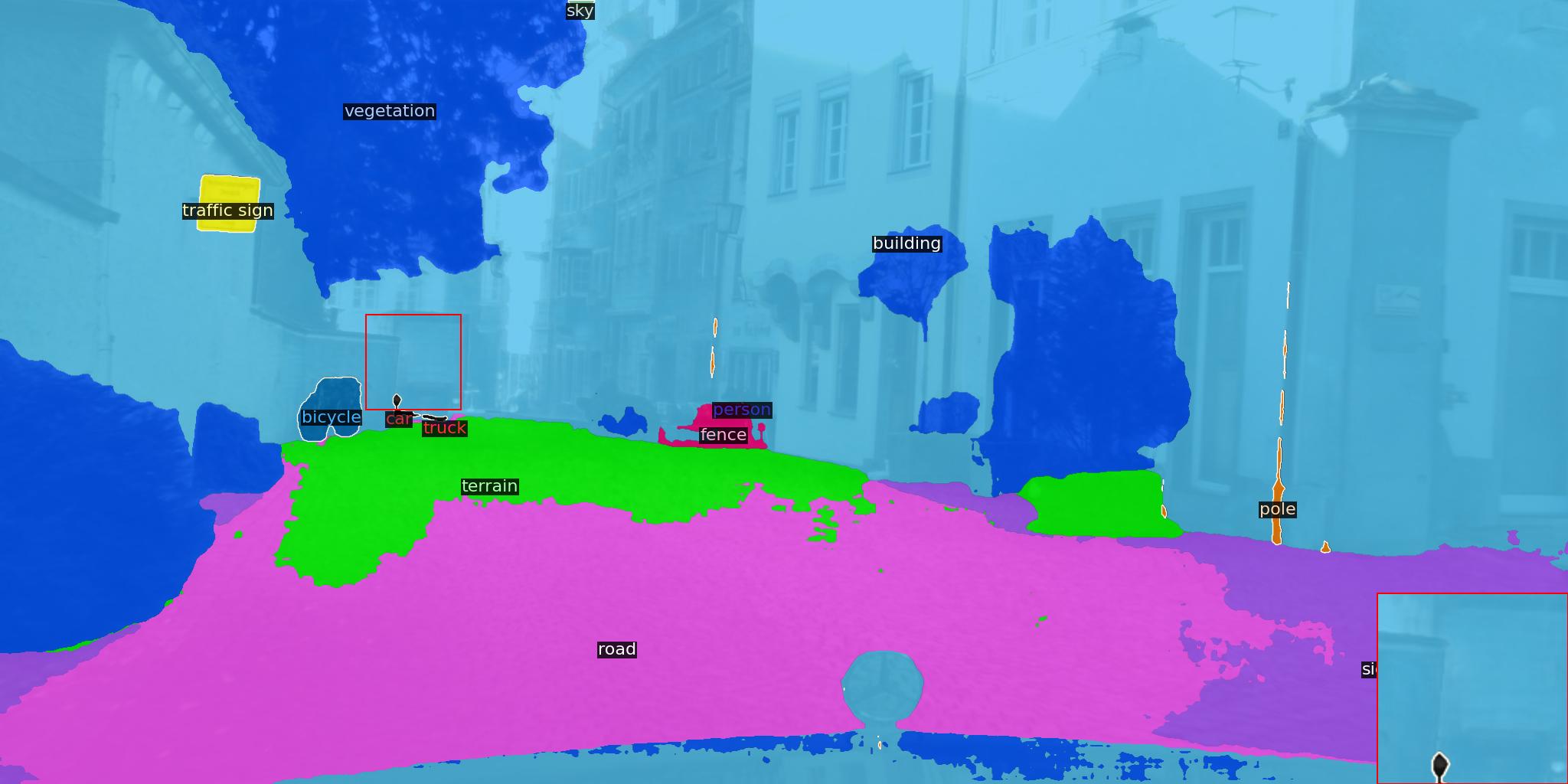}&\includegraphics[height=1.7cm, width=2cm]{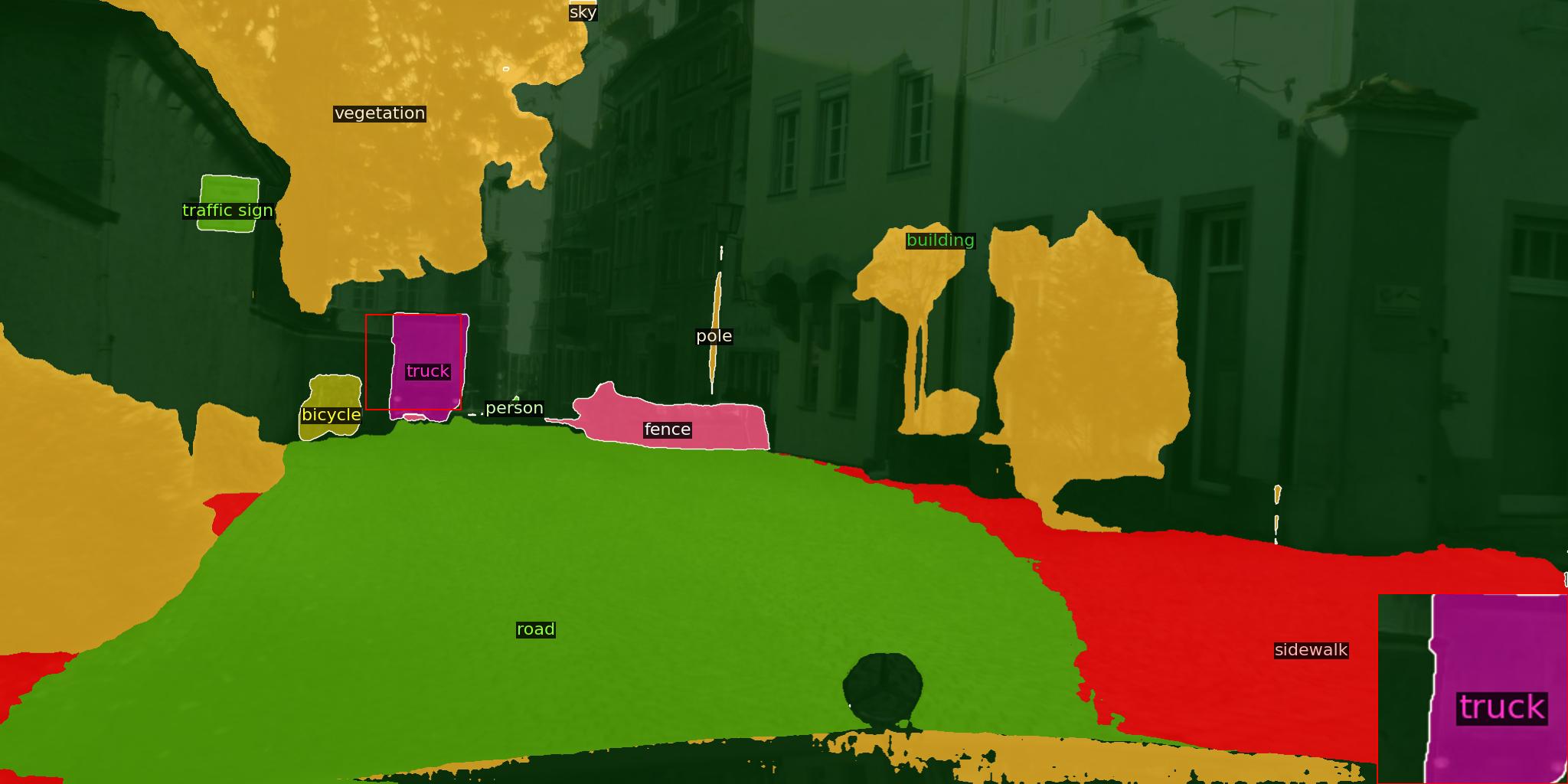}\\

    \end{tabular}
    \vskip-5pt         \caption{Predictions of Mask2Former~\cite{mask2former} on novel task of raindrop removal and pre-training task of defogging. After restoration using LoRA and LoRA-A, the results improve significantly. Furthermore, LoRA-A yields better segmentation results for the pre-training task of defogging.}
    \label{fig:seg}
\end{figure}

\begin{figure}
    \centering
    \setlength{\tabcolsep}{1pt}
    \small
    \begin{tabular}{ccccc}
         &Image&\makecell{Degraded\\ prediction}&LoRA&LoRA-A  \\

        \rotatebox[origin=c]{90}{Novel\hspace{-40pt}}&\includegraphics[height=1.7cm, width=2cm]{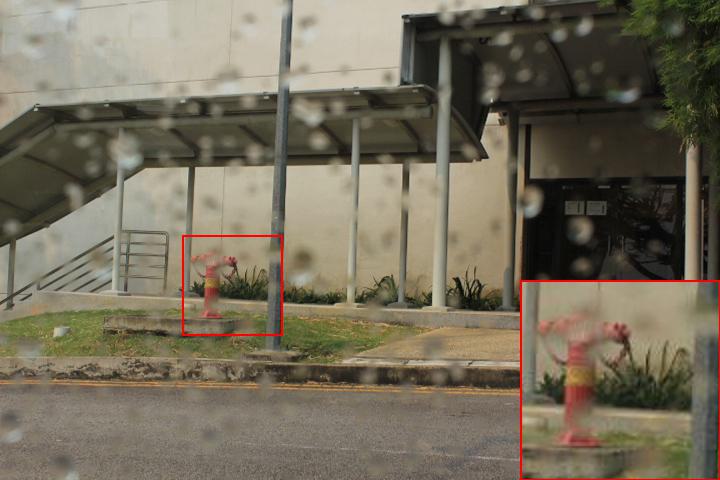}&\includegraphics[height=1.7cm, width=2cm]{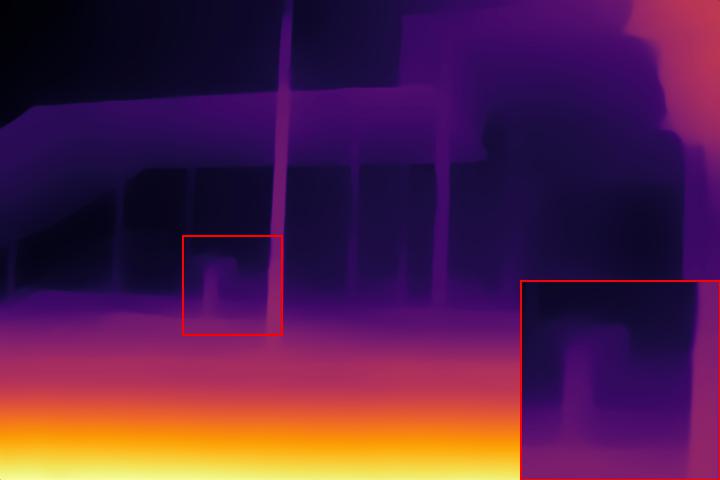}&\includegraphics[height=1.7cm, width=2cm]{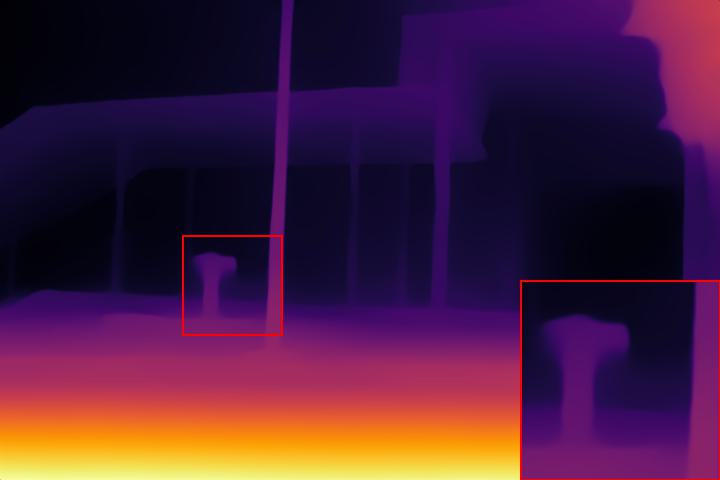}&\includegraphics[height=1.7cm, width=2cm]{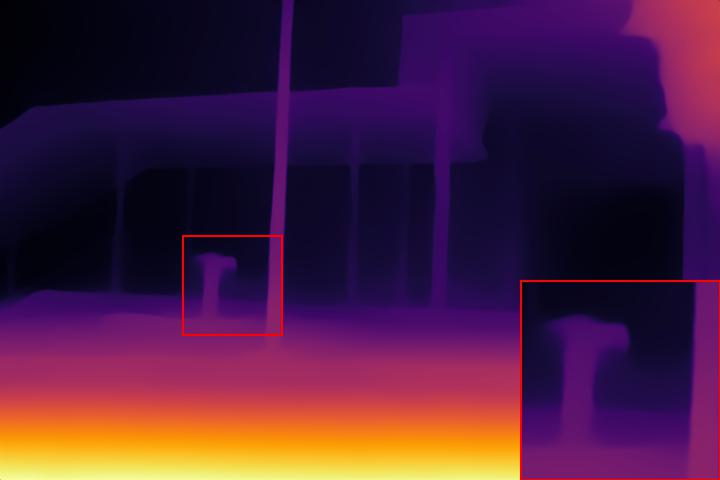}\\

         \rotatebox[origin=c]{90}{Pre-train\hspace{-45pt}}&\includegraphics[height=1.7cm, width=2cm]{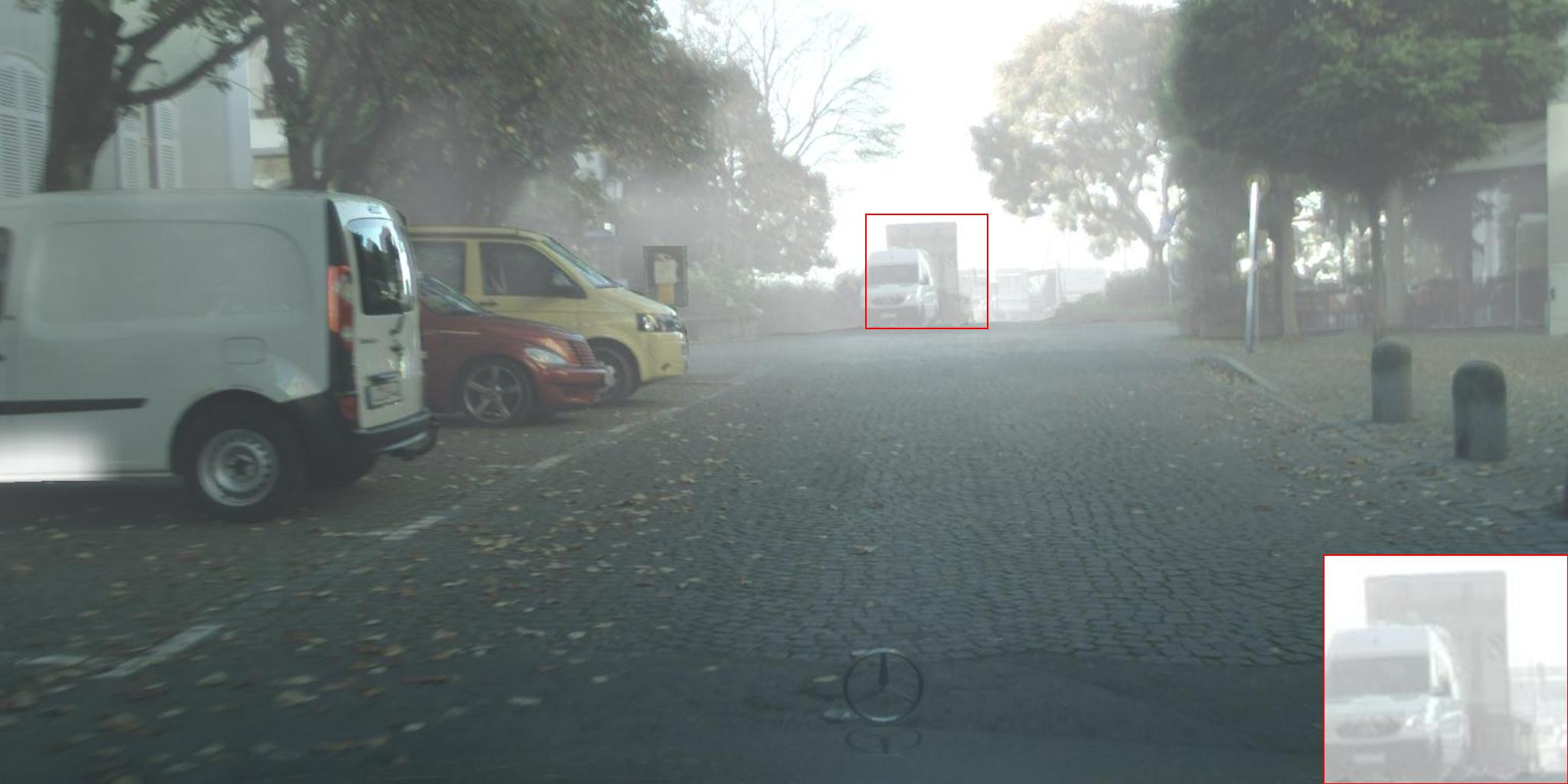}&
         \includegraphics[height=1.7cm, width=2cm]{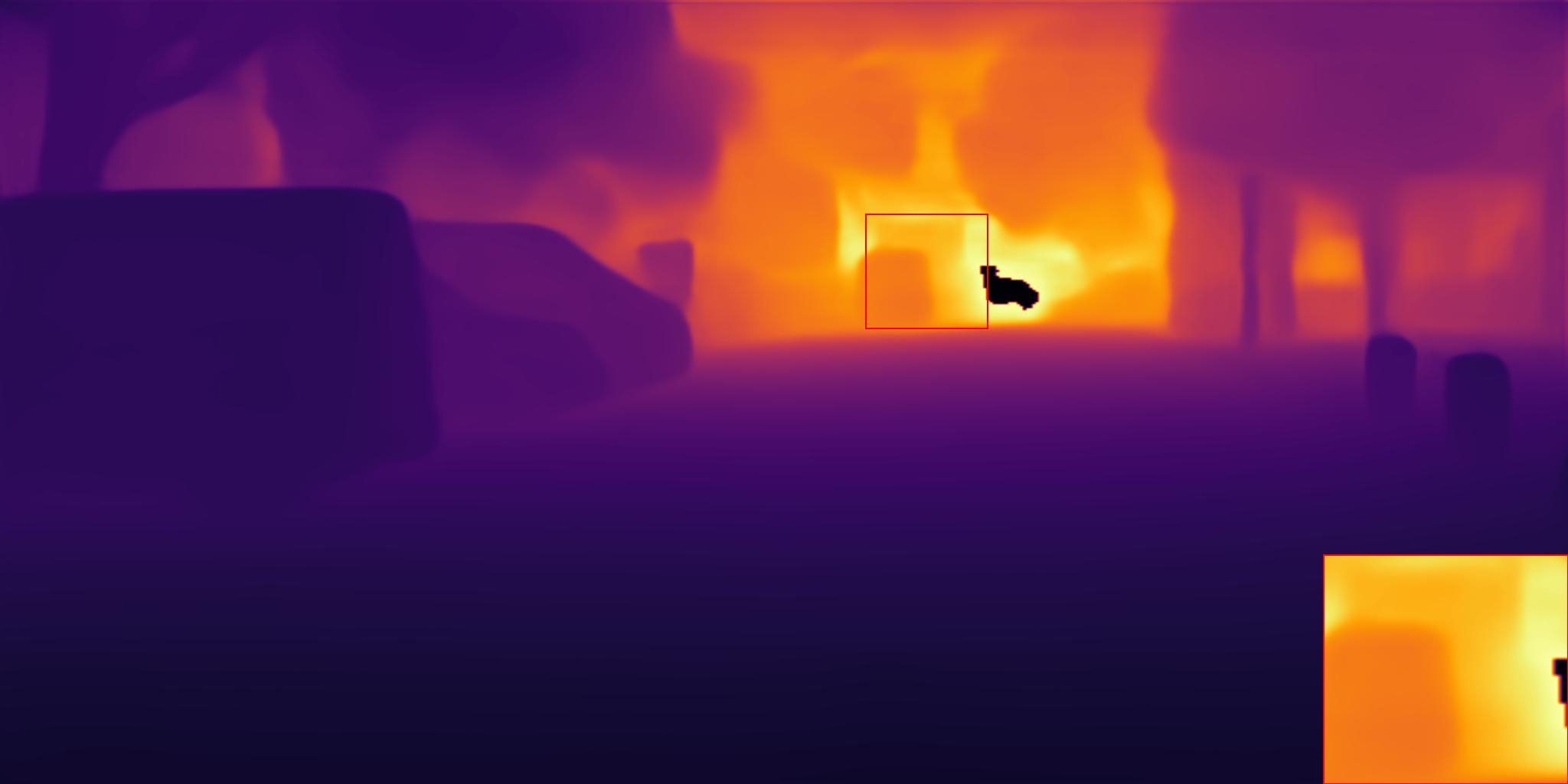}&
         \includegraphics[height=1.7cm, width=2cm]{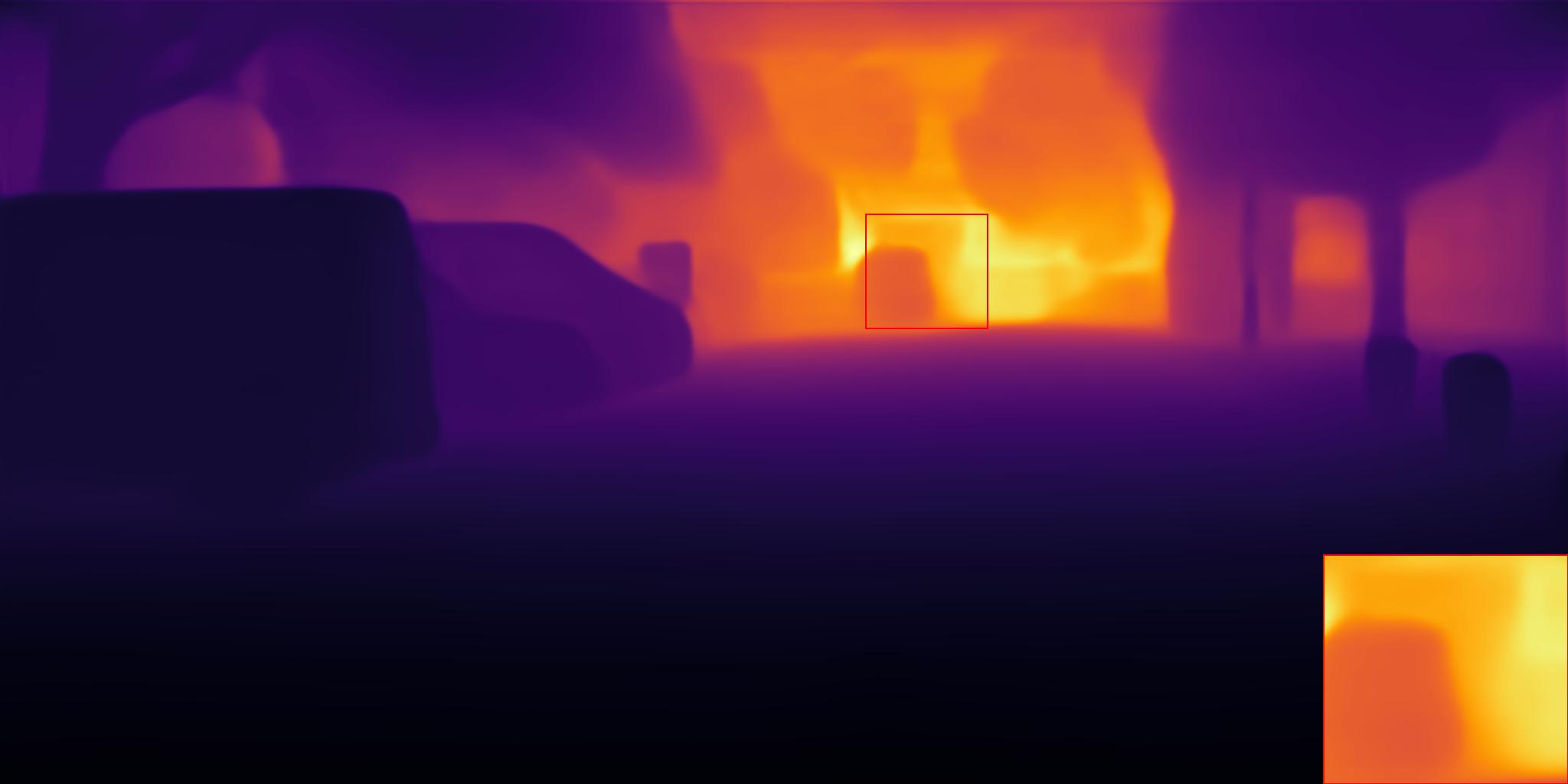}&
         \includegraphics[height=1.7cm, width=2cm]{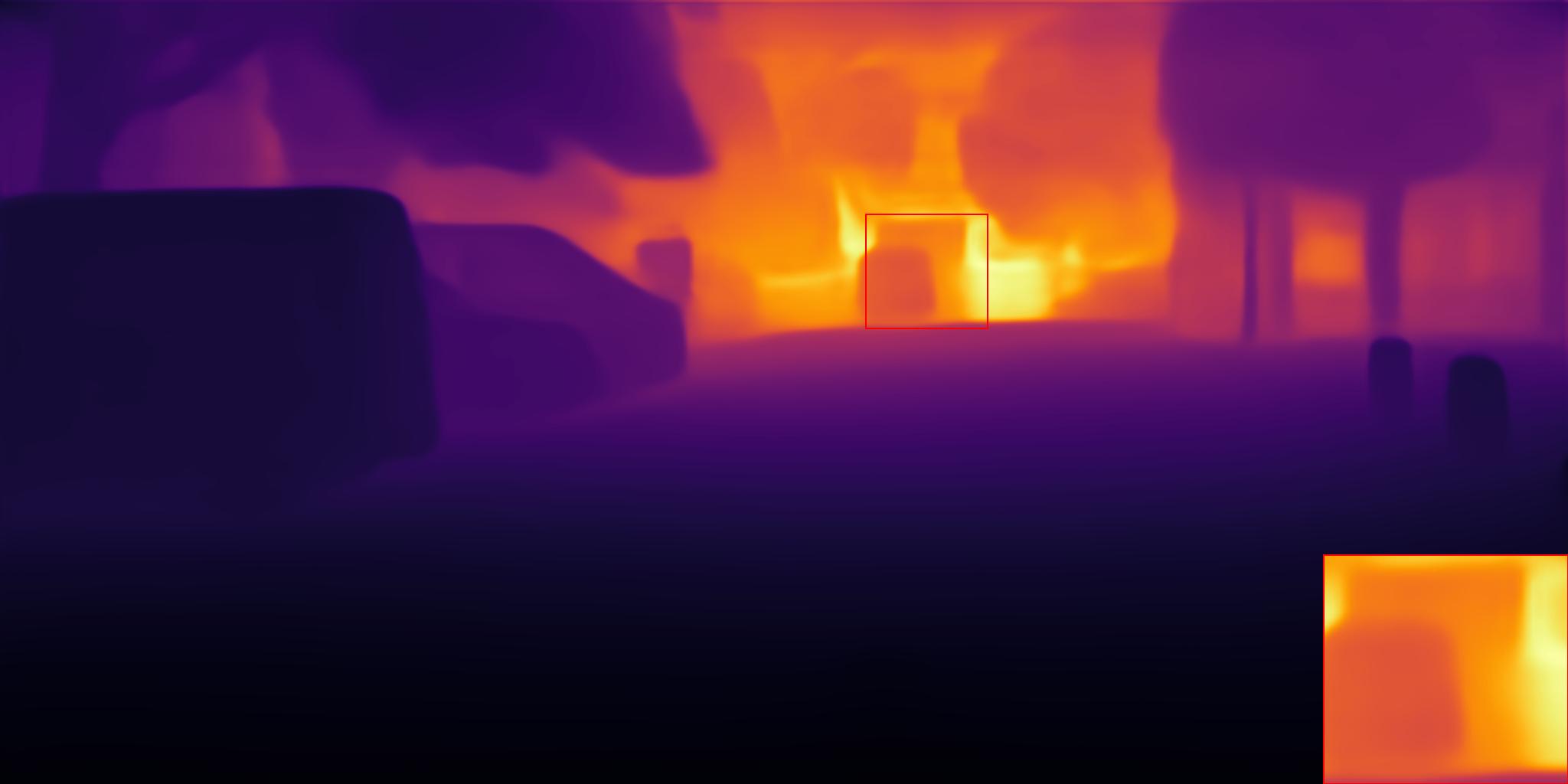}\\

    \end{tabular}
\vskip-5pt     \caption{Predictions of Depth-Anything~\cite{depthanything} on the novel task of raindrop removal and pre-training task of defogging. The predictions improve significantly after restoration using LoRA and LoRA-A. LoRA-A yields better depth results for the pre-training task of defogging.}
    \label{fig:depth}
\end{figure}

\begin{table}[t]
    \centering
    \setlength{\tabcolsep}{2pt}
    \caption{mIoU scores of Mask2Former on images restored by LoRA and LoRA-A adapted models for the pre-training task of defogging and novel task of raindrop removal.}
    \begin{tabular}{c|c|c|c|c}
         Task&Degradation&Degraded&LoRA&LoRA-A  \\ \hline
         Novel& Raindrop&39.19&43.82&42.25\\
         Pre-train&Fog&57.00&58.16&59.14
    \end{tabular}
    \label{tab:seg}
\end{table}


\begin{table}[t]
    \centering
    \setlength{\tabcolsep}{2pt}
    \caption{Quantitative evaluation of Depth Anything~\cite{depthanything} on images restored by LoRA and LoRA-A for defogging (pre-training task) and raindrop removal (novel task).}
    \begin{tabular}{c|c|c|c|c}
         Task, Degradation& $\delta_1\uparrow$ &$\delta_2\uparrow$&$\delta_3\uparrow$&SILOG$\downarrow$\\
         \hline
         Novel, Raindrop (Degraded)& 0.950& 0.986&0.995&0.137\\
         Novel, Raindrop (LoRA)&0.977&0.995&0.998&0.106\\
         Novel, Raindrop (LoRA-A)&0.974&0.995&0.998&0.112\\ \hline
         Pre-train, Fog (Degraded)&0.978&0.989&0.991&0.199\\ 
         Pre-train, Fog (LoRA)&0.980&0.991&0.992&0.190\\
         Pre-train, Fog (LoRA-A)&0.982&0.991&0.993&0.186\\
    \end{tabular}
    \label{tab:depth}
\end{table}

\noindent\textbf{Semantic segmentation. }We use Mask2Former~\cite{mask2former} to compare the predictions obtained on the degraded images and restored images produced by the adapted models (using LoRA and LoRA-A). The quantitative and qualitative results of this experiment are presented in Table~\ref{tab:seg} and Fig.~\ref{fig:seg}, respectively. We use the mean Intersection over Union (mIoU) to evaluate segmentation performance. For the novel task of raindrop removal, performance significantly improves after restoration using LoRA and LoRA-A. Note that the predictions of Mask2Former on the clean images were used as ground truth for raindrop removal as Raindrop~\cite{raindrop} does not contain segmentation annotations. For the pre-training task of defogging, LoRA-A preserves restoration performance thereby leading to better predictions of Mask2Former. Thus, the adapted models can greatly aid in improving the segmentation performance.\\

\noindent\textbf{Depth estimation. }For the task of depth estimation, we use a foundation model called Depth Anything~\cite{depthanything}. Table~\ref{tab:depth} and Fig.~\ref{fig:depth}, respectively provide the quantitative and qualitative comparisons of Depth Anything on the degraded image and the images restored using LoRA and LoRA-A. For quantitative evaluation we use $\delta_1, \delta_2, \delta_3$ thresholds and SILOG metrics (see~\cite{depthmetrics}). To calculate metrics for both raindrop removal and defogging, the predictions of Depth Anything on the clean images were used as ground truth. There is a significant improvement in depth estimation on the novel task after restoration using LoRA and LoRA-A. Furthermore, LoRA-A performs better on the pre-training tasks leading to superior depth predictions for defogging.



\section{Conclusions}
We proposed leveraging LoRA to efficiently adapt pre-trained AWIR models to novel restoration tasks, achieving near-identical performance to full fine-tuning but at a fraction of the learnable parameters. 
Additionally, we introduced LoRA-Align, which significantly recovers pre-training task performance with a minor trade-off on novel tasks while maintaining LoRA's parameter efficiency. 
Finally, we demonstrated that AWIR models adapted using LoRA and LoRA-A can enhance the performance of autonomous navigation tasks such as semantic segmentation and depth estimation for adverse weather conditions. We believe that our work will aid the real-world deployability of AWIR models.

\section{Acknowledgments}
This work is supported by the Intelligence Advanced Research Projects Activity (IARPA) via Department of Interior/ Interior Business Center (DOI/IBC) contract number 140D0423C0076. The U.S. Government is authorized to reproduce and distribute reprints for Governmental purposes notwithstanding any copyright annotation thereon. Disclaimer: The views and conclusions contained herein are those of the authors and should not be interpreted as necessarily representing the official policies or endorsements, either expressed or implied, of IARPA, DOI/IBC, or the U.S. Government.

\bibliographystyle{IEEEtran}
\bibliography{IEEEabrv,egbib}

\end{document}